\documentclass[letterpaper]{article} 
\usepackage{aaai25}  
\usepackage{times}  
\usepackage{helvet}  
\usepackage{courier}  
\usepackage[hyphens]{url}  
\usepackage{graphicx} 
\usepackage{natbib}  
\usepackage{caption} 
\usepackage{algorithm}

\usepackage{algorithmicx}
\usepackage[noEnd=true, indLines=false]{algpseudocodex}
\usepackage{amssymb}
\usepackage{amsmath}
\usepackage{latexsym}
\usepackage{amsfonts}
\usepackage{multirow} 

\usepackage{color}

\usepackage{newfloat}
\usepackage{listings}
\DeclareCaptionStyle{ruled}{labelfont=normalfont,labelsep=colon,strut=off} 
\floatstyle{ruled}
\newfloat{listing}{tb}{lst}{}
\floatname{listing}{Listing}
\title{CSR:Achieving 1 Bit Key-Value Cache via Sparse Representation}

\author {
    Hongxuan Zhang\textsuperscript{\rm 1}\footnote{The work conducted during the internship at Ant Group.},
    Yao Zhao\textsuperscript{\rm 2},
    Jiaqi Zheng\textsuperscript{\rm 1},
    Chenyi Zhuang\textsuperscript{\rm 2},
    Jinjie Gu\textsuperscript{\rm 2},
    Guihai Chen\textsuperscript{\rm 1}
}
\affiliations {
    \textsuperscript{\rm 1}Nanjing University\\
    \textsuperscript{\rm 2}Ant Group\\
    x\_zhang@smail.nju.edu.cn 
}

\usepackage{bibentry}
\begin{document}

\maketitle

\begin{abstract}

The emergence of long-context text applications utilizing large language models (LLMs) has presented significant scalability challenges, particularly in memory footprint. The linear growth of the Key-Value (KV) cache—responsible for storing attention keys and values to minimize redundant computations—can lead to substantial increases in memory consumption, potentially causing models to fail to serve with limited memory resources. To address this issue, we propose a novel approach called Cache Sparse Representation (CSR), which converts the KV cache by transforming the dense Key-Value cache tensor into sparse indexes and weights, offering a more memory-efficient representation during LLM inference. Furthermore, we introduce NeuralDict, a novel neural network-based method for automatically generating the dictionary used in our sparse representation. Our extensive experiments demonstrate that CSR achieves performance comparable to state-of-the-art KV cache quantization algorithms while maintaining robust functionality in memory-constrained environments. 



\end{abstract}

\section{Introduction}

The introduction of large language models (LLMs) has brought about a new wave of exciting AI applications, including document summarization, code analysis, extended multi-turn applications, tool learning, and more. Among these applications, those involving long text have garnered significant interest, such as RAG (Retrieval-Augmented Generation). RAG tackles the challenge of generating accurate and pertinent content, particularly in scenarios where queries extend beyond the training data or require up-to-date knowledge, by integrating external information sources. This fusion of RAG with LLMs expands the scope of LLMs and makes them increasingly applicable for specialized and knowledge-driven tasks in real-world contexts. However, the significant number of parameters in LLMs, amounting to tens or hundreds of billions, results in high memory and computation requirements during generation tasks, especially when handling long contexts like RAG. To effectively support large language models (LLMs), it is crucial to batch multiple requests together to minimize the cost per request.

\begin{figure}[ht]
    \centering
    \includegraphics[width=0.95\columnwidth]{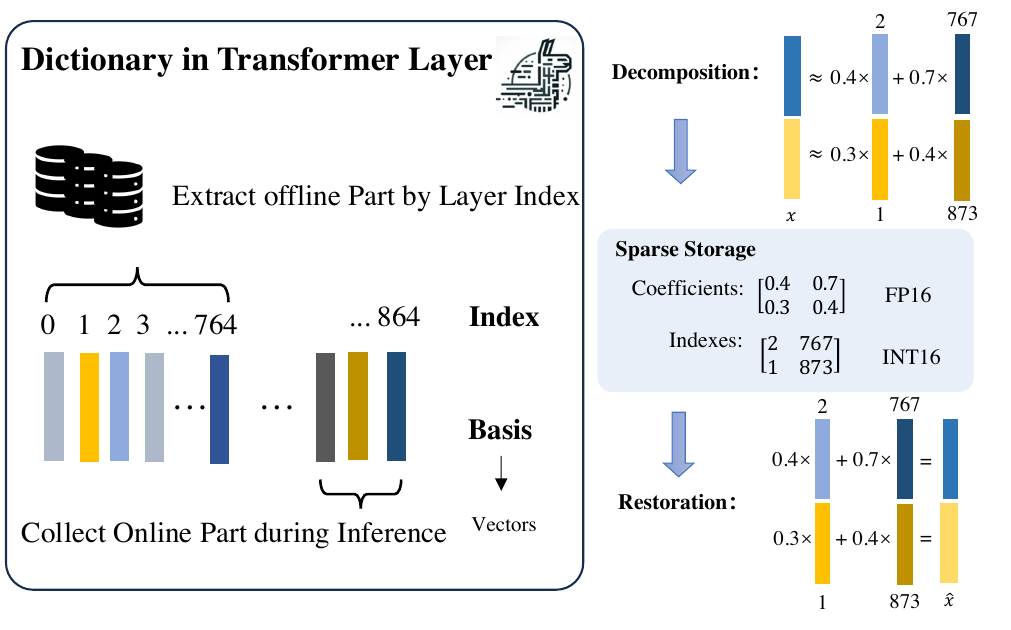}
    \caption{ The core of CSR is to use a dictionary that extracts the KV cache space feature to decompose dense original vectors into sparse coefficients and indexes in the dictionary, thereby significantly reducing the memory footprint required by the KV cache.}
    \label{fig:Introduction}
\end{figure}

The key-value (KV) cache utilized to store attention keys and values, and thereby avoid redundant computations, can lead to a substantial increase in memory usage and become a bottleneck for both speed and memory efficiency. The cache's memory consumption grows linearly with the length of the input prompt and the number of tokens generated, overshadowing even the substantial memory requirements of the model parameters. This presents scalability challenges for large language models, as the cache's linearly expanding memory footprint hinders its ability to handle longer sequences. Therefore, it is imperative to develop methods for compressing the KV cache to enable long-sequence inference in a memory-efficient manner.

We provide an overview of existing methods that help mitigate KV cache overhead as follows:\cite{Shazeer2019FastTD} introduces Multi-Query Attention, a variant of Multi-Head Attention(MHA)\cite{Vaswani2017AttentionIA}. MQA enables different heads to share the same KV caches, effectively reducing memory usage. Moreover, \cite{Ainslie2023GQATG} propose Grouped-Query Attention (GQA), offering a trade-off between performance degradation introduced by MQA and the memory footprint of MHA. These adjustments to the attention mechanism itself objectively reduce the memory footprint of the KV cache.
Another set of techniques utilizes quantization to reduce the number of bits used by the original stored data type, sacrificing data precision for memory footprint.
Additionally, some researchers have taken an alternative approach by lowering the memory footprint of the KV cache through the eviction of unimportant cache parts from the GPU. We will delve into a detailed discussion of these two methods in section \ref{sec:related work}

In this paper, we propose CSR (Cache Sparse Representation), which offers a sparse representation of the KV cache and provides an equivalent but less memory-intensive representation for the original KV cache during LLM inference. Our contributions are outlined as follows:

\begin{enumerate}
    \item CSR presents a novel solution for addressing the high memory footprint of the KV cache in long-text LLM applications. It is not only applicable to various existing attention mechanisms in transformers but also independent of well-established solutions such as KV cache quantization and KV cache eviction.
    \item Our extensive experiments on various models and datasets demonstrate that CSR not only delivers comparable performance to 4-bit or 2-bit KV cache quantization algorithms under relatively abundant memory conditions but also maintains robust performance with less than 1 bit per channel in memory-constrained situations.
\end{enumerate}

\section{Preliminary}


\subsection{Sparse representation}

Sparse representation is a well-researched fields in computer vision and pattern recognition. However, to the best of our knowledge, no work yet try using sparse representation to reduce the memory footprint used during large language model inference. Suppose we have a dictionary $D=[ \mathbf{d_1}, \mathbf{d_2}, ...,  \mathbf{d_N}] \in \mathbb{R}^{d\times N}$ and each \textit{basis} vector $\mathbf{d}_n$ in $D$ is an $l_2$-norm unity vector. Define the sparsity of a representation vector $\mathbf{r}$ as the $l_0$ norm of $\mathbf{r}$, which means the number of the nonzero elements of vector $\mathbf{r}$. Given dictionary $D$
and limit maximum representation sparsity as $s$, for a dense origin vector $\mathbf{x} \in \mathbb{R}^{d}$, the way to find the sparse representation of $\mathbf{x}$ is solving the following optimization problem:

\begin{equation}
\label{equ:sro problem}
\mathbf{r}( \mathbf{x}, D, s ) = \arg\min\| \mathbf{x} - D\mathbf{r} \|^2 \quad \text{s.t.} \quad \| \mathbf{r} \|_0 \leq s
\end{equation}

\noindent where $\mathbf{r} \in \mathbb{R}^{N}$ means sparse representation of $\mathbf{x}$ with sparsity no greater than $s$. Among different types of algorithms for solving equation (\ref{equ:sro problem}), Matching Pursuit(MP) \cite{mallat1993matching} is the earliest and widely used one to generate sparse representation satisfying sparsity limitations. The core idea of the MP is to iteratively choose the best atom from the dictionary based on a certain similarity measurement to approximately obtain the sparse solution. First of all, the residual vector is initialized as $\mathbf{R}_0=\mathbf{x}$, $\mathbf{r} = \mathbf{0} \in \mathbb{R}^{d}$. The MP algorithm will determine the optimal atom vector index $i_{g}$ and the corresponding coefficient $c_{g}$ through the following two formulas,

\begin{equation}
    \label{equ: cg}
    c_{g} = \mathop{sup}|\mathbf{R}_{g} \cdot \mathbf{d}_n|
\end{equation}
\begin{equation}
    \label{equ: ig}
    i_{g} = \mathop{\mathrm{argmax}}_{n}|\mathbf{R}_{g} \cdot \mathbf{d}_{n} |
\end{equation}
\noindent where $ 1 \leq g \leq s$ represents the number of current iterations. Subsequently, update $\mathbf{r}[i_{g}]=c_{g}$, and the residual vector $R_{g}$ is updated based on the part that was already approximated in the previous iteration as following:
\begin{equation}
    \label{equ: update_r}
    \mathbf{R}_{g+1} = \mathbf{R}_{g} - c_{g} \times \mathbf{d}_{i_g}
\end{equation}
\noindent MP will repeat calculating (\ref{equ: cg})-(\ref{equ: update_r}) until $c_{s}$ and $i_{s}$ are calculated and $\| \mathbf{r} \|_0 = s$ exactly. 


\subsection{KV cache in attention}

LLM inference can be divided into the prefill phase and the decoding phase. In the prefill phase, each token of the input prompt is used to generate a KV cache for every transformer layer of LLMs. The model uses and updates the KV cache to generate the next token autoregressively in the decoding phase. Since the KV cache mechanism for different attention heads is the same, we will not consider the attention head index in the subsequent discussion.

Assuming a model's hidden size is $d$ and the number of key (or value) attention heads is $h$, let $X^{\lambda}_{p} \in \mathbb{R}^{b\times l \times h \times d_h}$ represent the activations of the input prompt $p$'s tokens after being forwarded into transformer layer $\lambda$, where $b$ is batch size, $l$ is the length of prompt tokens, and $d_h = d//h $ is the tensor size for each attention head. $W^{\lambda}_K, W^{\lambda}_V \in \mathbb{R}^{d\times d}$ of the current layer will map $X^{\lambda}_{p}$ to key and value cache through the following equation:

\begin{equation}
    X^{\lambda}_{p, \{ K,V \} } = X^{\lambda}_{p} W^{\lambda}_{\{ K,V \}}
\end{equation}
Here, $\lambda$ is the transformer layer index. $X^{\lambda}_{p, K}$, $X^{\lambda}_{p,V}$ are cached in the memory as KV cache of prompt $p$ for layer $\lambda$ (Here we temporarily ignore the impact of position embedding). During the autoregressive decoding phase, each forward pass generates a new token $t$, and its corresponding activations after passing through layer $\lambda$ are represented as $X^{\lambda}_{t} \in \mathbb{R}^{b \times 1 \times d}$. After being mapped to the key (K) and value (V) space using $W^{\lambda}_K$ and $W^{\lambda}_V$, the corresponding $X^{\lambda}_{t,K}$ and $X^{\lambda}_{t,V}$ are appended to the KV cache of layer $\lambda$.  Throughout the remainder of the paper, we will use $X^{\lambda}_{\{K,V\}}$ to refer to the K or V cache space of the transformer layer $\lambda$.

\section{KV Cache Sparse Representation:CSR}


In this section, we introduce our method, Cache Sparse Representation (CSR), which utilizes the dictionary that fully extracts KV cache features and replaces dense KV cache vectors with sparse indexes and coefficients to significantly reduce the memory footprint during inference. We initially present our intuitions collected during the LLM inference stage, which directly guide the dictionary construction of CSR. Subsequently, we provide a comprehensive overview of the CSR procedure and delve into the detailed process of constructing the dictionary required by CSR.

\subsection{Intuitions}
\label{sec: obv}
We extracted a range of prompts from wikitext dataset\cite{merity2016pointer}, and forward them into Llama, a widely utilized public model. Subsequently, we gathered the KV cache generated during model inference. To aid in subsequent observation and research, we reduced the collected KV cache to a two-dimensional space through PCA in the channel dimension. This allowed us to derive the following observations through analysis.



\begin{figure}[htb!]
    \centering
    \includegraphics[width=\linewidth]{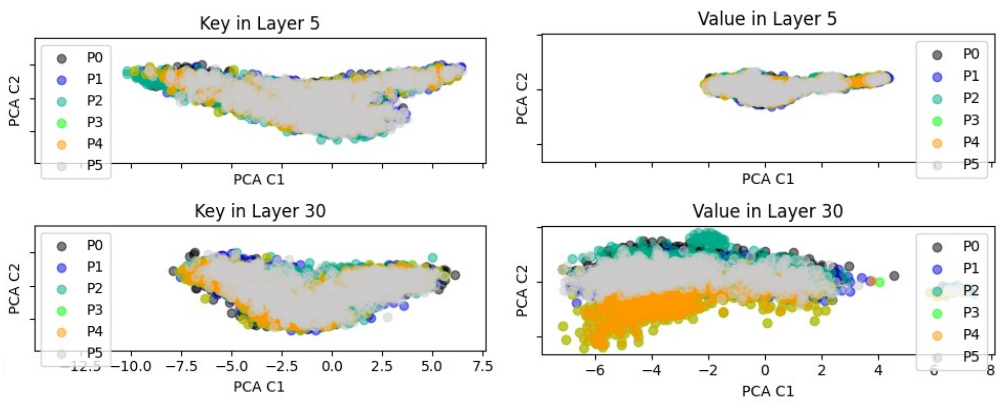}
    \caption{The distribution of $X_{K}$ among prompts is nearly identical. While there is substantial spatial overlap in $X_{V}$ in the shallow layer, few noticeable differences 
still emerge in the deep layers (e.g., Layer 30).}
    \label{fig:differentPrompt}
\end{figure}

\noindent\textbf{Difference among prompts is nearly ignorable}. Following PCA dimensionality reduction to a two-dimensional space, we observe that the spaces covered by different prompts are nearly identical, as depicted in Figure \ref{fig:differentPrompt}. This finding suggests that a portion of the constructed dictionary can be shared across different query prompts. We refer this query-independent part as the \textit{offline} part. Note that few noticeable differences still exist in the deep transformer layers. We propose the \textit{online} part to deal with this issue.


\begin{figure}[htb!]
    \centering
    \includegraphics[width=\columnwidth]{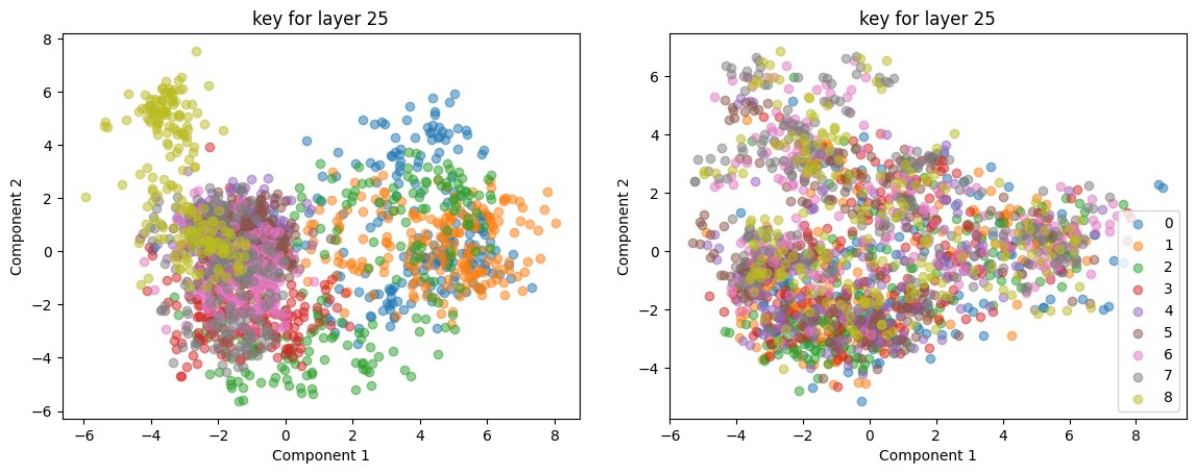}
    \caption{The key cache in layer 25 is evenly segmented into 8 groups based on their positions. It is evident that, in comparison with the keys processed by RoPE, the keys that have not undergone RoPE processing are thoroughly intermingled which is better for extracting basis vectors.}
    \label{fig:RopeOrNot}
\end{figure}

\begin{figure}[htb!]
    \centering
    \includegraphics[width=\columnwidth]{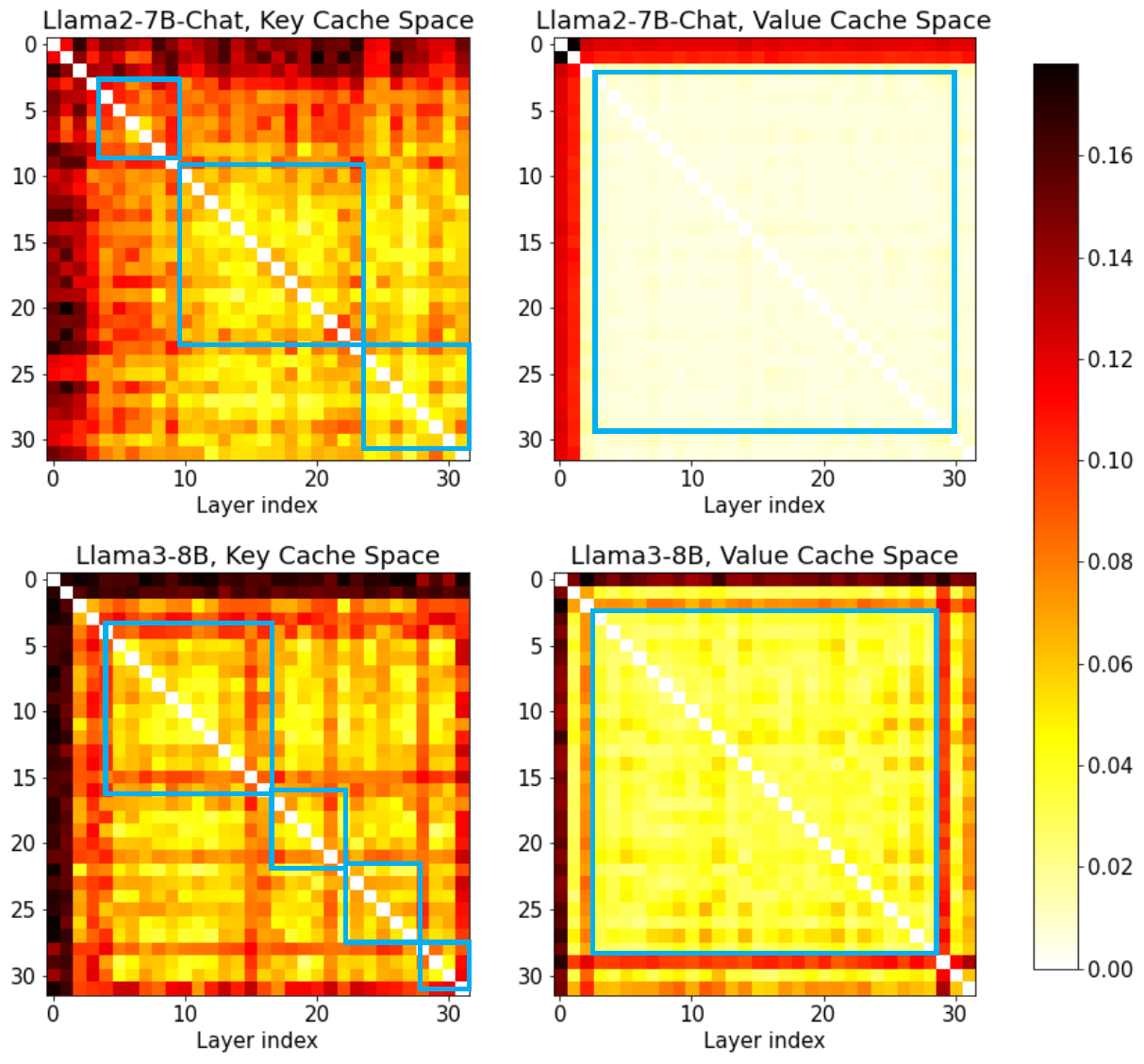}
    \caption{JS divergence for $X_{\{K,V\}}$ from different transformer layers. The lighter the color, the smaller the distribution difference.We use blue boxes to highlight adjacent layers with similar KV cache space.}
    \label{fig:js-div}
\end{figure}

\noindent\textbf{Position embedding makes nonstationary Keys}. An important consideration in determining the sparse representation for Keys is managing the positional embedding such like RoPE, which is applied to Keys and Queries in most public LLMs through embedding relative positional information between Keys and Queries.  The nature of RoPE causes the Keys to be relatively unstable with regard to position, as depicted in Figure \ref{fig:RopeOrNot}. Due to this phenomenon, we opt to pre-process the Key cache of tokens before introducing the position embedding.


\noindent\textbf{Adjacent transformer layers' KV space is similar}. To analyze the differences in $X^{\lambda}_{\{K,V\}}$ between Transformer layers, we first normalize the collected KV cache into $l_2$-norm unity vectors, then perform PCA in pairs on adjacent layers to reduce the dimension to 2. After that we generate a two-dimensional histogram of $200 \times 200$ bins to obtain the discrete distribution of $X_{\{K,V\}}$. Finally, we measure the difference of KV cache space from two transformer layers by calculating the JS divergence between these discrete distributions, part of the results are shown in the figure \ref{fig:js-div}. We observes that the distribution of $X_{\{K,V\}}$ between most adjacent layers is similar. So we decide to to construct a multi-layers shared offline dictionary based on the similarity between layers in order to save memory footprint as much as possible. Take $X_{K}$ as an example, the set of layers aggregated is denoted as $\mathcal{M}_{K} = \{ \Lambda_{1}, ... \Lambda_{i}  \}$, and $\Lambda_{i} = \{ \lambda_{m}, ... \lambda_{n} \} $. For $\forall i \neq j,  \Lambda_{i} \cap \Lambda_{j} = \varnothing $, and 
$\bigcup_{i} \Lambda_{i} $ is the set of all transformer layers. Any transformer layer pair $(\lambda_{m}, \lambda_{n})$ in the same $\Lambda_{i}$ satisfies:

\begin{equation}
\label{equ:pair_js}
JSD( P^{\lambda_{m}}_{K} \Vert P^{\lambda_{n}}_{K} ) \leq \delta_{1}, \forall \lambda_{m}, \lambda_{n} \in \Lambda_{i}
\end{equation}

\begin{equation}
\label{equ:seq_js_sum}
\sum_{\lambda_{m} \in \Lambda_{i} } JSD( P^{\lambda_{m}}_{K} \Vert P^{\lambda_{m+1}}_{K} ) \leq \delta_{2}, with \quad \lambda_{m+1} \in \Lambda_{i} 
\end{equation}

\noindent where $P^{\lambda_{m}}_{K}$ represents the discrete distribution obtained through dimensional histogram after reducing the dimensions of the vector in the attention head to 2 dimensions using PCA, $\delta_1$ and $\delta_2$ are thresholds. Equation \ref{equ:pair_js} serves to limit the similarity of the cache space distribution of any two transformer layers in the $\Lambda_{i}$ , while equation \ref{equ:seq_js_sum} is utilized to prevent the cache space after aggregation from becoming excessively large.

Based on these observations, we propose the following guidelines for constructing the dictionary needed for CSR: 
\begin{itemize}
    \item First, the construction of the dictionary will be divided into two parts: offline and online. 
    \item Second, we choose to preprocess $X_K$ prior to the embedding of positional information.
    \item Third, we fuse the $X_{\{K,V\}}$ from adjacent transformer layers to construct a multi-layers shared offline dictionary based on the similarity between layers in order to save memory footprint as much as possible.
\end{itemize}


\begin{algorithm}[tb!]
\caption{NeuralDict}
\label{alg:TrainND}
\begin{algorithmic}[1] 
    \Procedure{NeurDict}{$\mathcal{C}, m, N, E$} 
    \State \textbf{Input:} The calibration corpus dataset $\mathcal{C}$, language model $m$, offline dictionary size $N$ and training procedure epochs number $E$.
    \State Perform inference on dataset $\mathcal{C}$ using model $m$ and collect $X^{m}_{K}, X^{m}_{V}$ for each layer and attention head in model $m$
    \State Generate $\mathcal{M}_{K}$ and $\mathcal{M}_{V}$ based on Equation \ref{equ:pair_js} and \ref{equ:seq_js_sum} for Key and Value respectively.

    \State \Call{TrainOnMergedLayers}{$\mathcal{M}_{K}, N, X, E$}
    \State \Call{TrainOnMergedLayers}{$\mathcal{M}_{V}, N, X, E$}
    
    \State return all trained neural weight as offline dictionary.
    \EndProcedure

    \Statex

    \Procedure{TrainOnMergedLayers}{ $\mathcal{M}, N, X, E$}
    \For{$\Lambda_{i} \in \mathcal{M}$}
        \State $X$=concatenate [$X^{\lambda_n}$ for $\lambda_n \in \Lambda_{i}$]
         \State \Call{TrainNeurDict}{$N, X, E$}
    \EndFor
    \EndProcedure
    
    \Statex
  
    \Procedure{TrainNeurDict}{$N, X_{\{K,V\}}, e$} 
    \State \textbf{Input:} Offline dictionary size $N$, and Key cache or Value cache $X_{\{K,V\}}$ in corpus dataset, epochs $e$ to train
    \State Initialize $W_{D}=[d_{1}, \ldots, d_{N}]$ with cluster centroids of $X_{\{K,V\}}$.
    \LComment{Normalize the vector to ensure its magnitude is 1}
    \State $W_{D}$ = \Call{ReNorm}{$W_{D}$}
    
    \For{$\hat{e} = [1, \ldots, e]$}
    \For{Batch $\mathcal{B}\in X^{l,h}_{\{K,V\}} $}
       \State calculate $\mathcal{L}$ using equation \ref{equ:completeLoss}
       \State backward $\mathcal{L}$ and update $W_{D}$
        \State update adaptive coefficient $\beta$
        \State $W_{D}$ = \Call{ReNorm}{$W_{D}$}
    \EndFor
    \EndFor
    \EndProcedure

    \Statex
    \Procedure{ReNorm}{$W_{D}$}
    \For{i in [1,2,\ldots, N]}
        \State $\widehat{w}_i = \frac{w_i}{\|w_i\|_2}$
    \EndFor
    \State $\widehat{W} = [\widehat{w}_1 ; \widehat{w}_2;\ldots;\widehat{w}_N]$
    \State return $\widehat{W}$
    \EndProcedure 
    
\end{algorithmic}
\end{algorithm}

\subsection{CSR's Workflow}

We have divided CSR into two stages. In the Preparation Stage, for a language model, CSR probes $X_{\{K,V\}}$ of each transformer layer using the calibration corpus dataset, then aggregates $X_{\{K,V\}}$ of each layer based on the JS divergence of the discrete distribution, and trains to obtain an offline dictionary that can be shared by multiple layers. The other stage is the Inference Stage, in which CSR replaces the original KV cache of the language model and utilizes the sparse representations to reduce the GPU memory footprint.


\subsection{Preparation Stage}
\label{sec: dict_cons}
\label{sec:offline-dict}


The primary concern is how to construct a dictionary that can approximately represent each KV cache tensor in LLM generated by the current query by selecting only $s$ bases in the dictionary. Clustering is a widely used unsupervised learning method for extracting features from a vector space. However, the clustering algorithm does not directly interact with the process of calculating the sparse representation in CSR. As a result, the dictionary constructed by clustering does not take into consideration the features of residual tensors beyond the first iteration in MP. To address these issues, we propose a novel neural network-based method named \textit{NeuralDict} to automatically resolve this problem.




\subsubsection{NeuralDict} The offline dictionary construction remains consistent across $\mathcal{M}_{K} $ or $\mathcal{M}_{V}$ of the model. We utilize the calibration set $\mathcal{C}$ as the corpus dataset to assess the distribution of $X_{\{K,V\}}$ in each layer of the large language model $m$. For a model $m$ with hidden states size in each attention head as $d_h$, split into $s_n$ chunks, and given the dictionary $D$ with a size of $N$, the dictionary we aim to create can be viewed as a matrix $W_{D} \in \mathbb{R}^{d_h // s_n \times N}$. This matrix $W_{D}$ can be considered as the learnable weights in a single linear layer neural network without any bias or activation function. We utilize the mean squared error as shown in Equation \ref{equ: MSE} to train $W_{D}$. Take Key cache as example:

\begin{equation}
    \label{equ: MSE}
    \mathcal{L}_{MSE} = \sum_{x \in X_K} \| \mathbf{x} - W_{D} \mathbf{r}(\mathbf{x}, W_{D}, s  ) \|_{2}^2 
\end{equation}
where $ \mathbf{r}(\mathbf{x}, W_{D}, s )$ represents the sparse representation vector calculated by the MP algorithm, and $W_D$ serves as the basis vector's dictionary. In practical applications, we set $s=8$ to strike a balance between training effectiveness and efficiency. If $s$ is too small, the mean squared error $\mathcal{L}_{MSE}$ will be excessively large and difficult to decrease, while a large $s$ will result in a prolonged MP process, leading to lower training efficiency. After updating $W_D$ through loss backpropagation, we apply an additional update to $W_{D}$ as:
\begin{equation}
W_{D} = \text{ReNorm}(W_{D})
\end{equation}
\noindent where $\text{ReNorm}$ denotes the normalization of each column vector in $\mathbb{R}^{d*N}$ to $l_2$ norm unity as shown in Algorithm \ref{alg:TrainND}.

\noindent\textbf{Adaptive Regularization to Encourage Diversity}. To prevent the training from getting trapped in local optima due to similarities between pairs of $d_{n} \in D$ during training, we include the following regularization term in the training function to promote the diversity of vectors in $W_{D}$:
\begin{equation}
 \mathcal{L}_{div} = \frac{ 1 }{ N^{2} }  \| I -  W_{D}^T W_{D}  \|_{F}^2
\end{equation}
Here, $F$ denotes the Frobenius norm, and $I \in \mathbb{R}^{N \times N} $ represents the identity matrix. In practical training, we observed that the $X_{\{K,V\}}$ space corresponding to the shallow transformer layer is simpler than that of the deep layer, leading to a very small $\mathcal{L}_{MSE}$ for this part. Since the magnitude of the mean squared error (MSE) loss varies with the transformer layer while the diversity term does not, we incorporate an adaptive coefficient to adjust the weight of $ \mathcal{L}_{MSE}$ and $\mathcal{L}_{div}$:

\begin{equation}
    \label{equ:completeLoss}
    \mathcal{L} = \mathcal{L}_{MSE} + \beta \mathcal{L}_{div}
\end{equation}
where $\beta = min(0.1 \times \frac{ \hat{\mathcal{L}}_{MSE}}{ \hat{\mathcal{L}}_{div}}, 1.0)$. Note that $\hat{\mathcal{L}}_{MSE}$ and $\hat{\mathcal{L}}_{div}$ represent the calculated values of the last batch without any gradient information. The purpose of limiting $\beta$ to 1.0 is to prevent the model from overly focusing on reducing $\mathcal{L}_{div}$ and disregarding $\mathcal{L}_{MSE}$ when $\mathcal{L}_{div}$ is sufficiently small. The whole training procedure is shown in Algorithm \ref{alg:TrainND}.





    






\subsection{Inference Stage}

When the language model's transformer layer is loaded into the GPU, CSR will load the layer's corresponding offline dictionary onto the same device. Note that due to the existence of Merged Layers, we prefer to load layers corresponding to the same offline dictionary onto one device. The whole process of how CSR take place of original KV cache is illustrated in Figure \ref{fig:Introduction}.

\subsubsection{Build dictionary}

For a new prompt $p$, CSR build $D_{K}^{\lambda}(p)$ and $D_{V}^{\lambda}(p)$ as dictionaries for the Key cache and Value cache correspondingly. For each transformer layer $\lambda$. CSR will extract the corresponding part from the offline dictionary for the transformer layer according to the layer index as show in Figure \ref{fig:Introduction}. In addition to the offline part, the dictionary also has an online part obtained by performing random sampling and reverse sampling from the calculated KV cache. In order to prevent poor fitting results caused by out of distribution of some KV cache during inference, we follow the KV quantization framework such as \cite{kang2024gear} and design a separate part for outlier entries. 



\subsubsection{KV Decomposition and Sparse Storage}

CSR will compute the sparse representation for the tokens in the prompt using $D_{K}^{\lambda}(q)$ or $D_{V}^{\lambda}(q)$ for the $X_{ {K,V} }$ by solving problem \ref{equ:sro problem} using the Matching Pursuit algorithm. The maximum sparsity is set to be $s$ which is so-called \textit{MP-level}, the Matching Pursuit algorithm will perform $s$ iterations to generate sparse representations with a sparsity of $s$ for the entire $X_{\{K,V\}}$. We denote the sparse representations of the KV cache as $\mathbf{r}( X_{\{K,V\}}, D_{\{K,V\}}^{\lambda}(q), s)$ with sparsity $s$ in layer $\lambda$. Please note that there are no more than $s$ non-zero elements in $\mathbf{r}$.Therefore, it is only necessary to store the index and coefficient of these non-zero elements. The index indicates the position of the selected basis vector in the dictionary, while the value represents the corresponding coefficient. 

\subsubsection{De-Sparse to restore}
To meet the needs of calculating attention scores,  CSR will de-sparse $\mathbf{r}$ into a tensor form similar to the original KV cache:
\begin{equation}
\tilde{X}^{\lambda}_{ \{ K,V \} } = D^{\lambda}_{\{ K,V \}} \mathbf{r}( X^{\lambda}_{\{K,V\}}, D^{\lambda}_{\{ K,V \}}, s)
\end{equation}
Here, $\tilde{X}_{ \{ K,V \} } \in \mathbb{R}^{b\times l_g \times h \times d_h}$, and $l_g$ represents the number of prompt tokens and generated tokens. $\tilde{X}_{ \{ K,V \} }$ will be used in the attention score calculation instead of original KV cache. When a new token is generated, the KV cache corresponding to the new token will also be replaced by CSR with a sparse representation during the subsequent inference process.

\subsubsection{Analysis for CSR}

Each attention head's initial $X_{K,V}$ comprises $d_h$ floating-point values, with fp16 being the prevalent datatype in LLM inference. Under CSR, just $s\times s_n$ fp16 values for coefficients, accompanied by $s \times s_n$ INT16 values for indexes. The compression rate can be calculated as $\frac{16 \times d_{h}}{16 s \times s_n + 16s \times s_n} = \frac{d_h}{2s\times s_n}$, which implies that for CSR($s, s_n$), the number of bits of the corresponding quantization algorithm is $\frac{16}{d_h/2s \times s_n} = \frac{32s\times s_n}{d_h} $ bits. Taking LLaMA3-8B as an example, it has $d_h$ =128. For CSR($s=4, s_n=1$), the corresponding quantization bit count is 1 bit.



\section{Experiments}

\subsection{Experiments Settings}

\noindent\textbf{Models} We have applied CSR to multiple LLMs using the HuggingFace transformers' codebase. To assess CSR's effectiveness across various attention mechanisms, we conducted experiments on the Llama2-7B-chat\cite{touvron2023llama, touvron2023llama1}, Llama3-8B-Instruct\cite{llama3modelcard} and Baichuan2-7B-chat\cite{baichuan2023baichuan2}. Among them, Llama2-7B-chat and Baichuan2-7B-chat use Multi-Head Attention, while Llama3-8B-Instruct uses Grouped-Query Attention.

\noindent\textbf{Benchmark} The primary goal of CSR is to reduce the memory usage of the KV cache by identifying sparse representations for the KV cache within a long context setting. To evaluate its effectiveness, we utilized the LongBench benchmark \cite{bai2023longbench}, which is a bilingual and multitask benchmark designed to assess the long context understanding capabilities of LLM. In our evaluation, we relied on standard metrics such as F1 score, ROUGE score, and similarity score. These metrics align with the settings established in \cite{Liu2024KIVIAT} for different datasets within the LongBench.

\noindent\textbf{CSR} In the experimental section, all value caches use the result of $s_n=2$. For simplicity, we use CSR-$s$ to refer to the MP-level of size s. Please note that for the Value cache, since $s_n=2$, this means that the maximum MP-level it corresponds to is only half of that of the Key cache. For example, CSR-8 means that for the Key cache, $s=8, s_n=1$, but for the Value cache, it is $s=4, s_n=2$.

\noindent\textbf{Baselines} We selected state-of-the-art (SOTA) KV cache quantization algorithms to establish a robust baseline for measuring CSR performance. These included:
\begin{itemize}
    \item KIVI\cite{Liu2024KIVIAT}: KIVI proposed quantizing the Key cache per-channel and the Value cache per-token. To achieve this, KIVI introduced a tuning-free quantization algorithm known as KIVI-2, supporting 2 bits, and KIVI-4, supporting 4 bits.
    \item GEAR\cite{kang2024gear}: GEAR applies 4-bit quantization to the majority of entries in the KV cache and utilizes a low-rank matrix to approximate the quantization error. Additionally, GEAR uses a sparse matrix to handle outliers.
\end{itemize}

\noindent\textbf{Hardware Environment}
A single NVIDIA A100 GPU (80GB) with 128GB memory.




\subsection{Robust Performance on various tasks}

\begin{table*}[htb]

\begin{tabular}{l|l|llllllllllll}
\hline
\textbf{Model}                                                             & \textbf{Method} & \textbf{\begin{tabular}[c]{@{}l@{}}2wiki-\\ mqa\end{tabular}} & \textbf{\begin{tabular}[c]{@{}l@{}}hotp-\\ otqa\end{tabular}} & \textbf{\begin{tabular}[c]{@{}l@{}}musi-\\ que\end{tabular}} & \textbf{trec}  & \textbf{\begin{tabular}[c]{@{}l@{}}narr-\\ ative-\\ qa\end{tabular}} & \textbf{\begin{tabular}[c]{@{}l@{}}qas-\\ per\end{tabular}} & \textbf{\begin{tabular}[c]{@{}l@{}}qm-\\ sum\end{tabular}} & \textbf{lcc}   & \textbf{\begin{tabular}[c]{@{}l@{}}sam-\\ sum\end{tabular}} & \textbf{\begin{tabular}[c]{@{}l@{}}trivi-\\ aqa\end{tabular}} & \textbf{\begin{tabular}[c]{@{}l@{}}multi-\\ field-\\ qa\_en\end{tabular}} & \textbf{Avg}   \\ \hline
\multirow{8}{*}{\begin{tabular}[c]{@{}l@{}}Llama2-\\ 7B-Chat\end{tabular}} & FP16            & 26.47                                                         & 33.84                                                         & 9.33                                                         & 68.14          & 16.65                                                                & 17.17                                                       & 20.81                                                      & 40.88          & 58.25                                                       & 82.74                                                         & 35.62                                                                     & 37.26          \\ \cline{2-14} 
                                                                           & GEAR            & 26.52                                                         & 32.65                                                         & 9.01                                                         & 68.17          & 16.78                                                                & 18.03                                                       & 21.12                                                      & 57.59          & 41.66                                                       & 83.53                                                         & 36.65                                                                     & 37.42          \\
                                                                           & KIVI-4          & 26.08                                                         & 33.48                                                         & 9.53                                                         & 68.14          & 17.14                                                                & 17.16                                                       & 20.61                                                      & 58.07          & 40.33                                                       & 82.49                                                         & 36.06                                                                     & 37.19          \\
                                                                           & CSR-16          & \textbf{27.04}                                                & \textbf{33.67}                                                & \textbf{9.72}                                                & \textbf{68.17} & \textbf{17.49}                                                       & 15.39                                                       & 21.08                                                      & \textbf{58.78} & 40.69                                                       & 83.46                                                         & \textbf{36.65}                                                            & \textbf{37.58} \\ \cline{2-14} 
                                                                           & KIVI-2          & 26.06                                                         & 32.13                                                         & 9.71                                                         & 68.14          & 16.69                                                                & 19.30                                                       & 20.79                                                      & 57.21          & 39.31                                                       & 83.23                                                         & 36.32                                                                     & 37.17          \\
                                                                           & CSR-8           & 25.55                                                         & 32.27                                                         & 9.05                                                         & 68.10          & \textbf{17.16}                                                       & 15.30                                                       & 20.75                                                      & \textbf{57.76} & 38.52                                                       & \textbf{83.72}                                                & 34.42                                                                     & 36.60          \\ \cline{2-14} 
                                                                           & CSR-6           & 26.47                                                         & 32.72                                                         & 9.23                                                         & 68.17          & 16.91                                                                & 14.94                                                       & 20.7                                                       & 38.22          & 56.71                                                       & 83.22                                                         & 33.12                                                                     & 36.40          \\ \cline{2-14} 
                                                                           & CSR-4           & 24.75                                                         & 30.24                                                         & 8.74                                                         & 64.36          & 15.59                                                                & 12.54                                                       & 19.91                                                      & 35.02          & 43.82                                                       & 81.18                                                         & 30.06                                                                     & 33.29          \\ \hline
\end{tabular}
\caption{We conducted experiments on CSR methods employing varying $s$ and corresponding quantization methods with an identical number of bits. We highlight the data where our method performs better within the same group. As there is no equivalent method for quantization below 2 bits, we present CSR-4 corresponding to 1 bit and CSR-6 corresponding to 1.5 bit.}
\label{tab:main}

\end{table*}

\begin{table*}[htb]
\begin{tabular}{l|l|llllllllllll}
\hline
\textbf{Model}                                                                     & \textbf{Method} & \textbf{\begin{tabular}[c]{@{}l@{}}2wiki-\\ mqa\end{tabular}} & \textbf{\begin{tabular}[c]{@{}l@{}}hotp-\\ otqa\end{tabular}} & \textbf{\begin{tabular}[c]{@{}l@{}}musi-\\ que\end{tabular}} & \textbf{trec} & \textbf{\begin{tabular}[c]{@{}l@{}}narr-\\ ative-\\ qa\end{tabular}} & \textbf{\begin{tabular}[c]{@{}l@{}}qas-\\ per\end{tabular}} & \textbf{\begin{tabular}[c]{@{}l@{}}qm-\\ sum\end{tabular}} & \textbf{lcc} & \textbf{\begin{tabular}[c]{@{}l@{}}sam-\\ sum\end{tabular}} & \textbf{\begin{tabular}[c]{@{}l@{}}trivi-\\ aqa\end{tabular}} & \textbf{\begin{tabular}[c]{@{}l@{}}multi-\\ field-\\ qa\_en\end{tabular}} & \textbf{Avg} \\ \hline
\multirow{5}{*}{\begin{tabular}[c]{@{}l@{}}Llama3-\\ 8B-Ins-\\ truct\end{tabular}} & FP16            & 32.17                                                         & 33.23                                                         & 17.87                                                        & 74.54         & 20.24                                                                & 29.22                                                       & 22.83                                                      & 42.63        & 56.49                                                       & 88.79                                                         & 38.81                                                                     & 41.53        \\
                                                                                   & CSR-16          & 32.08                                                         & 34.89                                                         & 17.98                                                        & 74.19         & 20.29                                                                & 28.56                                                       & 22.33                                                      & 40.97        & 58.3                                                        & 89.39                                                         & 39.41                                                                     & 41.67        \\
                                                                                   & CSR-8           & 30.06                                                         & 34.11                                                         & 16.27                                                        & 71.63         & 20.18                                                                & 23.44                                                       & 21.72                                                      & 38.92        & 56.45                                                       & 89.43                                                         & 37.49                                                                     & 39.97        \\
                                                                                   & CSR-6           & 29.56                                                         & 35.57                                                         & 15.45                                                        & 71.26         & 19.34                                                                & 20.02                                                       & 21.53                                                      & 38.52        & 55.49                                                       & 89.33                                                         & 34.35                                                                     & 39.13        \\
                                                                                   & CSR-4           & 27.76                                                         & 33.84                                                         & 15.47                                                        & 71.22         & 19.68                                                                & 17.23                                                       & 21.46                                                      & 37.9         & 54.75                                                       & 88.62                                                         & 29.67                                                                     & 37.96        \\ \hline
\multirow{5}{*}{\begin{tabular}[c]{@{}l@{}}Baichu-\\ an-7B\end{tabular}}           & FP16            & 20.13                                                         & 26.52                                                         & 11.51                                                        & 73.46         & 17.66                                                                & 21.02                                                       & 22.04                                                      & 17.05        & 63.53                                                       & 81.02                                                         & 39.78                                                                     & 35.79        \\
                                                                                   & CSR-16          & 19.9                                                          & 25.49                                                         & 11.52                                                        & 72.95         & 18.73                                                                & 18.83                                                       & 21.9                                                       & 17.03        & 59.63                                                       & 80.94                                                         & 38.11                                                                     & 35.00        \\
                                                                                   & CSR-8           & 18.7                                                          & 25.34                                                         & 10.95                                                        & 72.38         & 18.06                                                                & 18.08                                                       & 21.46                                                      & 17.3         & 56.91                                                       & 80.11                                                         & 34.09                                                                     & 33.94        \\
                                                                                   & CSR-6           & 20.06                                                         & 25.42                                                         & 10.28                                                        & 71.95         & 16.68                                                                & 16.94                                                       & 21.06                                                      & 15.94        & 56.64                                                       & 79.58                                                         & 33.41                                                                     & 33.45        \\
                                                                                   & CSR-4           & 19.04                                                         & 23.65                                                         & 9.87                                                         & 71.17         & 15.92                                                                & 16.12                                                       & 21.21                                                      & 15.31        & 54.14                                                       & 77.85                                                         & 31.63                                                                     & 32.35      \\ \hline
\end{tabular}
\caption{We conducted experiments on Longbench using Llama3-8B and Baichuan2-7B, and the results showed that CSR is 
also effective for these models.}
\label{tab: various models}
\end{table*}


Initially, we present a comparison between CSR and various quantization algorithms on the Llama2-7B-chat model. For KIVI and GEAR, we perform grid search on hyperparameters and show the results of the best obtained. The hidden size of each attention head in the Llama2-7B-chat model is 128 so according to the previous analysis, CSR-8 is equivalent to 2 bits in quantization, and CSR-16 is equivalent to 4 bits in quantization. We grouped several methods according to equivalent quantization levels, namely FP16 corresponding to 16 bits, GEAR, KIVI-4 and CSR-16 corresponding to 4 bits, and KIVI-2 and CSR-8 corresponding to 2 bits in Table \ref{tab:main}. The performance of various methods on multiple datasets, is presented in Table \ref{tab:main}. For the 4-bit group, our method performs better than KIVI and GEAR on most datasets, while for the 2-bit group, our method and KIVI have their own advantages and disadvantages. We conclude that CSR, KIVI, and GEAR exhibit similar performances and CSR can provide performance comparable to state-of-the-art 4-bit or 2-bit quantization algorithms.

\noindent\textbf{Effective CSR with Less Than 2 bit:} There is no way to reduce from 2bit to 1bit for quantization based methods. However, CSR can provide sparse representation for all KV caches at less than 2 bits or even 1 bit per channel, thus alleviating the tight memory resources of the GPU without any KV cache eviction. We conducted extensive experiments with CSR-6, equivalent to 1.5 bit, and CSR-4, equivalent to only 1 bit. In this scenario, CSR can still maintain performance on most datasets, with only a slight performance drop as shown in Table \ref{tab:main}. Even CSR-4 only drops 8\% in model performance compared to FP16, but the memory occupied by KV cache is less than $\frac{1}{10}$.

\subsection{CSR works well for various language models}

CSR is independent of the attention mechanism utilized in LLM, making it theoretically applicable to various models. In order to validate the versatility of our method across different models, we conducted more experiments on Baichuan2-7B, Llama3-8B-Instruct. As depicted in Table \ref{tab: various models}. The results demonstrate that despite providing at least an 8x compression ratio compared to the original data type, CSR still delivers strong performance across all models.

\subsection{Memory foorpint}

\begin{figure}
    \centering
    \includegraphics[width=0.95\columnwidth]{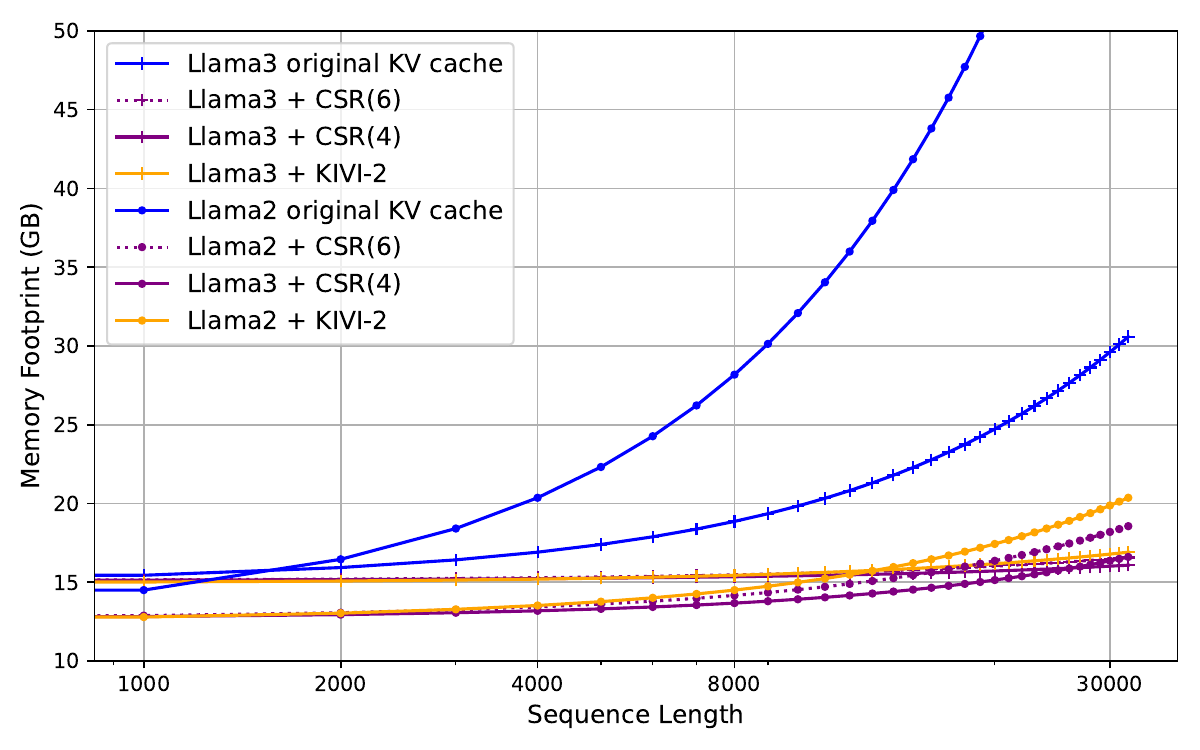}
    \caption{The figure is based on the Llama2-7B-chat and Llama3-8B-Instruct models, and shows the memory footprint used when using different methods for inference with batch size = 4. The x-axis is sequence length in log-scale, and the y-axis is the occupied memory.}
    \label{fig:mem-footprint}
\end{figure}

We plotted the relationship between the inference length and memory footprint of different models with KV cache using different methods as shown in Figure\ref{fig:mem-footprint}. As shown in the figure, the additional memory overhead introduced by the offline or online dictionary is almost negligible. Compared with the original KV cache, both CSR and quantization algorithms have greatly reduced the memory occupied by the KV cache. Compared with quantization, which cannot be further reduced from 2 bits, CSR provides the possibility of further reducing memory usage in long text scenarios, which means that larger inference lengths can be provided on GPUs with smaller memory.





\subsection{Effect of $s_n$ and NeuralDict size}

We analyze the role of $s_n$ based on the training results on NeuralDict. We choose the MSE loss of the Llama2-7B-chat model on all Merged layers, and the results are shown in Table \ref{tab: sn_key} and Table \ref{tab: sn_value}. Overall, from the perspective of Key cache and Value cache, the improvement of splitting the Value cache is very obvious, but there is almost no improvement when doing the Key cache. Based on this, in actual use, we fix $s_n=1$ for the Key cache in CSR.

Based on the results of Value Cache, we continue to analyze in depth the impact of the increase in $s_n$ on MSE loss. For s=4, the improvement brought by increasing $s_n$ from 2 to 4 is very significant, and when s=8, the significant improvement occurs when $s_n$ is increased from 1 to 2.

\begin{minipage}[bt!]{0.42\columnwidth}
    \centering
    \begin{tabular}{l|lll}
    \hline
    size                   & $s$ & $s_n$ & Loss   \\ \hline
    \multirow{6}{*}{8192}  & 4   & 1     & 0.259  \\
                           & 4   & 2     & 0.253  \\
                           & 4   & 4     & 0.093  \\ \cline{2-4} 
                           & 8   & 1     & 0.213        \\
                           & 8   & 2     & 0.094   \\
                           & 8   & 4     & 0.012   \\ \hline
    \multirow{6}{*}{16384} & 4   & 1     & 0.263   \\
                           & 4   & 2     & 0.223   \\
                           & 4   & 4     & 0.078   \\ \cline{2-4} 
                           & 8   & 1     & 0.194   \\
                           & 8   & 2     & 0.079   \\
                           & 8   & 4     & 0.008   \\ \hline
    \end{tabular}
    \label{tab: sn_value}
    \captionof{table}{NeuralDict for Value Cache}
\end{minipage}%
\hspace{0.1\columnwidth} 
\begin{minipage}[ht!]{0.42\columnwidth}
    \centering
    \begin{tabular}{l|lll}
    \hline
    size                   & $s$ & $s_n$ & Loss    \\ \hline
    8192                   & 4   & 1     & 0.079   \\
                           & 4   & 2     & 0.068   \\
                           & 4   & 4     & 0.052   \\ \cline{2-4} 
                           & 8   & 1     & 0.055   \\
                           & 8   & 2     & 0.047   \\
                           & 8   & 4     & 0.037   \\ \hline
    \multirow{6}{*}{16384} & 4   & 1     & 0.077   \\
                           & 4   & 2     & 0.058   \\
                           & 4   & 4     & 0.042   \\ \cline{2-4} 
                           & 8   & 1     & 0.047   \\
                           & 8   & 2     & 0.043   \\
                           & 8   & 4     & 0.033   \\ \hline
    \end{tabular}
    \label{tab: sn_key}
    \captionof{table}{NeuralDict for Key Cache}
\end{minipage}







\section{Related Work}
\label{sec:related work}

\noindent\textbf{KV Cache Quantization} Quantization is an alternative method for reducing memory and compute requirements during generation tasks, particularly in processing extremely long contexts. Prior research, such as that by \cite{Hooper2024KVQuantT1, Yue2024WKVQuantQW}, has focused on quantizing the KV cache. Meanwhile, \cite{Liu2024KIVIAT} proposes quantizing the key cache per-channel and the value cache per-token, \cite{kang2024gear} propose to use SVD to reduce the quantization error. However, these approaches are not applicable when the per-token quantization falls below 2 bits.

\noindent\textbf{KV Cache Eviction} Various approaches exist to minimize the KV cache footprint, with the common objective of retaining only a small subset of keys and values. One technique utilizes the attention mechanism's localized pattern, namely the attention sink, as proposed by \cite{Xiao2023EfficientSL}. This involves employing a finite attention window to retain only the "sink" token and a fixed number of recent tokens. Another strategy involves implementing a KV cache eviction policy considering the attention mechanism's sparsity. For example, \cite{Zhang2023H2OHO, ge2023model} suggest discarding non-essential parts of the KV cache to reduce memory usage during large language model (LLM) inference. Moreover, \cite{Liu2023ScissorhandsET} identifies a repetitive attention pattern during inference processes, recommending the retention of only "pivotal" tokens. Additionally, \cite{Anagnostidis2023DynamicCP} employs a learnable mechanism to identify uninformative tokens, implementing adaptive sparse attention that requires fine-tuning on the pre-trained model.


\section{Conclusion}
This paper introduces CSR, a framework for optimizing the memory footprint of the KV cache during LLM inference, based on compressed sensing algorithms. Our experiments on widely-used LLMs and long-context datasets have demonstrated that CSR's performance comparable to quantized algorithms when memory resources are relatively abundant (in comparison to 2-bit or 4-bit KV cache quantized algorithms). Furthermore, CSR exhibits robust performance even when memory is more constrained, aiming for less than 2 bits per channel. Notably, even with a per-channel bit count as low as 1, CSR can maintain robust performance. We believe that CSR provides an alternative approach for compressing the KV cache independently of quantization-related algorithms. Compared to quantization, CSR can operate effectively with a smaller quantization bit number, and maintain strong performance across the various tasks even with extremely low memory usage by KV cache.

\section{Limitations}

Compared with the quantization algorithm, CSR further reduces the memory occupied by the KV cache. However, the process of detecting the KV cache space of the model through the calibration dataset and then obtaining a part of the dictionary through offline training is time-consuming. We leave the research for a more efficient way to obtain the offline dictionary as future exploration.









\newpage



\bibliography{aaai25}

\newpage
\section{Appendix}

\subsection{Merged Layers for NeuralDict}

The thresholds for Equation (6) and Equation (7) is 0.20 and 1 respectively. The three models we experimented with, Llama2-7B-chat, Baichuan2-7B and Llama3-8B-Instruct, are all models with 32 transformer layers. There are slight differences in their merged results, but in order to facilitate the training of NeuralDict, we choose to make slight adjustments to the aggregated results. The adjusted results are as follows: The result for Key Cache is [ [0], [1], [2], [3,4,5], [6,7,8,9], [10,11,12,13], [14,15,16,17], [18,19,20,21], [22,23,24,25], [26,27,28,29], [30,31] ], and the result for Value Cache is [[0], [1], [2,3,4,5,6,7], [8,  9,  10, 11, 12, 13], [14, 15, 16, 17, 18, 19],  [20, 21, 22, 23, 24, 25], [26, 27, 28, 29, 30, 31]].

\begin{table}[h!]
\centering
\begin{tabular}{l|ll}
\hline
Size & Loss on Key & Loss on Value \\ \hline
1024 & 0.1023      & 0.1935        \\
2048 & 0.0662      & 0.1358        \\
4096 & 0.0585      & 0.1131        \\
8192 & 0.0539      & 0.1003        \\ \hline
\end{tabular}
\caption{Under different offline NeuralDict sizes, the MSE loss after convergence}
\label{tab: various offline size}
\end{table}

For key cache, $s=8$ and $s_n=1$ during training. As for value cache, $s=4$ and $s_n=2$. 
Different models have different choices for dictionary size. We tested the effect of different dictionary sizes on the final MSE loss on Llama3-8B-Instruct shown in Table \ref{tab: various offline size}. When the offline dictionary size increases from 1024 to 2048, there is a significant MSE loss decrease in both Key Cache and Value Cache. In addition, the decrease in loss caused by continuing to increase the offline size is not obvious. On Llama3-8B-Instruct, when the offline dictionary size is 2048, the size of each KV attention head is 2048/8=256. On Llama2-7B-chat and Baichuan, we use the same size setting, that is, 32*256=8192. According to the results shown in Experiments, we found that such an offline dictionary size setting has performed well enough.

\subsection{Effect of $\mathcal{L}_{div}$ term}

\begin{table}[h]
\begin{tabular}{l|l|lll}
\hline
Model                      & $\mathcal{L}_{div}$ & Epoch 10 & Epoch 20 & Converge \\ \hline
\multirow{2}{*}{llama2-7B} & w/     & 0.0628   & 0.0592   & 0.0561      \\
                           & w/o    & 0.0653   & 0.0593   & 0.0572      \\ \hline
\multirow{2}{*}{llama3-8B} & w/     & 0.0758   & 0.0694   & 0.0662      \\
                           & w/o    & 0.0815   & 0.0725   & 0.0702      \\ \hline
\end{tabular}
\caption{We calculated the mean value over all Merged Layers to evaluate the impact of $\mathcal{L}_{div}$ term on MSE loss.}
\label{tab: Abl_div}
\end{table}


We monitored the process of training the neural dictionary on all transformer layers and attention heads for the Llama-7B-chat and Llama3-8B-Insturct, and depicted the validation loss in Table \ref{tab: Abl_div}. It is evident that the presence of $\mathcal{L}_{div}$ results in a faster and more stable decline in $\mathcal{L}_{MSE}$ during training. Furthermore, the loss function value after convergence is also reduced.

\subsection{The impact of online part size on performance}

\begin{table*}[ht]
\begin{tabular}{l|l|llllllllllll}
\hline
\textbf{Method} & \textbf{\begin{tabular}[c]{@{}l@{}}Online-\\ Size\end{tabular}} & \textbf{\begin{tabular}[c]{@{}l@{}}2wiki-\\ mqa\end{tabular}} & \textbf{\begin{tabular}[c]{@{}l@{}}hotp-\\ otqa\end{tabular}} & \textbf{\begin{tabular}[c]{@{}l@{}}musi-\\ que\end{tabular}} & \textbf{trec} & \textbf{\begin{tabular}[c]{@{}l@{}}narr-\\ ative-\\ qa\end{tabular}} & \textbf{\begin{tabular}[c]{@{}l@{}}qas-\\ per\end{tabular}} & \textbf{\begin{tabular}[c]{@{}l@{}}qm-\\ sum\end{tabular}} & \textbf{lcc} & \textbf{\begin{tabular}[c]{@{}l@{}}sam-\\ sum\end{tabular}} & \textbf{\begin{tabular}[c]{@{}l@{}}trivi-\\ aqa\end{tabular}} & \textbf{\begin{tabular}[c]{@{}l@{}}multi-\\ field-\\ qa\_en\end{tabular}} & \textbf{Avg} \\ \hline
CSR-16          & 0                                                               & 26.28                                                         & 33.67                                                         & 9.72                                                         & 68.17         & 17.49                                                                & 15.39                                                       & 20.81                                                      & 56.78        & 38.98                                                       & 83.63                                                         & 36.65                                                                     & 37.12        \\
                & 8192                                                            & 27.04                                                         & 33.67                                                         & 9.72                                                         & 68.17         & 17.49                                                                & 15.39                                                       & 21.08                                                      & 58.78        & 40.69                                                       & 83.64                                                         & 36.65                                                                     & 37.58        \\
                & 16384                                                           & 27.04                                                         & 33.67                                                         & 9.72                                                         & 68.17         & 17.49                                                                & 17.12                                                       & 21.08                                                      & 58.28        & 40.88                                                       & 83.64                                                         & 36.65                                                                     & 37.61        \\ \hline
CSR-8           & 0                                                               & 23.93                                                         & 32.27                                                         & 7.69                                                         & 68.00         & 18.43                                                                & 14.09                                                       & 20.67                                                      & 37.64        & 48.39                                                       & 82.9                                                          & 33.36                                                                     & 35.21        \\
                & 8192                                                            & 25.55                                                         & 32.27                                                         & 9.05                                                         & 68.10         & 17.16                                                                & 15.30                                                       & 20.75                                                      & 57.76        & 38.52                                                       & 83.72                                                         & 34.42                                                                     & 36.60        \\
                & 16384                                                           & 25.86                                                         & 32.6                                                          & 9.12                                                         & 68.17         & {\color[HTML]{333333} 17.23}                                         & 15.65                                                       & 21.01                                                      & 38.63        & 57.22                                                       & 83.63                                                         & 35.12                                                                     & 36.74        \\ \hline
\end{tabular}
\caption{We conducted experiments on Longbench using Llama2-7B-chat with various online size, and the results showed that CSR is really robust to the size of online dictionary.}
\label{tab: online size}
\end{table*}

In Experiments section, the size of the online collection part used by CSR is set to be the same as the offline part. Specifically, the online collection size of each layer is 8192 for Llama2-7B-chat and Baichuan2-7B, while it is 2048 for Llama3-8B-Instruct. 

In order to study the impact of online size on performance, we conducted experiments based on $s=8$ and $s=16$ to test the effect of removing online size. The results are shown in the Table \ref{tab: online size}. On CSR-16, increasing the online size from 0 to 8192 has only a slight improvement, and only on the 2wikimqa, qmsum, lcc and samsum datasets. The overall performance improvement brought by increasing from 8192 to 16384 is even smaller, and the improvement mainly occurs on the qasper dataset. For CSR-8, the performance is slightly different. The performance improvement brought by increasing the online-size from 0 to 8192 is more obvious, while the improvement from 8192 to 16384 is still small. We conclude that an online size of 8192 is sufficient for CSR, and for smaller cases of $s$, the benefits brought by the online part are more obvious, and for $s=16$, the offline dictionary is sufficient to handle most cases, and the online part is almost not needed.

\end{document}